\documentclass[Afour,SageV,times]{sagej}

\usepackage{moreverb,url}

\usepackage[colorlinks,bookmarksopen,bookmarksnumbered,citecolor=red,urlcolor=red]{hyperref}

\newcommand\BibTeX{{\rmfamily B\kern-.05em \textsc{i\kern-.025em b}\kern-.08em
T\kern-.1667em\lower.7ex\hbox{E}\kern-.125emX}}

\def\volumeyear{2023}

\begin{document}

\runninghead{Morton et al.}

\title{Autonomous Material Composite Morphing Wing}

\author{Daniel Morton\affilnum{1,2}, Artemis Xu\affilnum{1}, Alberto Matute\affilnum{1}, Robert F. Shepherd\affilnum{1}}

\affiliation{\affilnum{1}Department of Mechanical and Aerospace Engineering, Cornell University, 124 Hoy Road, Ithaca, NY 14850, USA\\
\affilnum{2}Department of Mechanical Engineering, Stanford University, 440 Escondido Mall, Stanford, CA 94305, USA}

\corrauth{Daniel Morton}

\email{dpm263@cornell.edu}

\begin{abstract}
Aeronautics research has continually sought to achieve the adaptability and morphing performance of avian wings, but in practice, wings of all scales continue to use the same hinged control-surface embodiment. Recent research into compliant and bio-inspired mechanisms for morphing wings and control surfaces has indicated promising results, though often these are mechanically complex, or limited in the number of degrees-of-freedom (DOF) they can control. Seeking to improve on these limitations, we apply a new paradigm denoted Autonomous Material Composites to the design of avian-scale morphing wings. With this methodology, we reduce the need for complex actuation and mechanisms, and allow for three-dimensional placement of stretchable fiber optic strain gauges (Optical Lace) throughout the metamaterial structure. This structure centers around elastomeric conformal lattices, and by applying functionally-graded warping and thickening to this lattice, we allow for local tailoring of the compliance properties to fit the desired morphing. As a result, the wing achieves high-deformation morphing in three DOF: twist, camber, and extension/compression, with these morphed shapes effectively modifying the aerodynamic performance of the wing, as demonstrated in low-Reynolds wind tunnel testing. Our sensors also successfully demonstrate differentiable trends across all degrees of morphing, enabling the future state estimation and control of this wing. 
\end{abstract}

\keywords{Autonomous material composites, morphing wings, soft robotics, compliant lattice structures, metamaterials, additive manufacturing, optical strain sensing}

\maketitle

\begin{figure*}[t]
    \includegraphics[width=\textwidth]{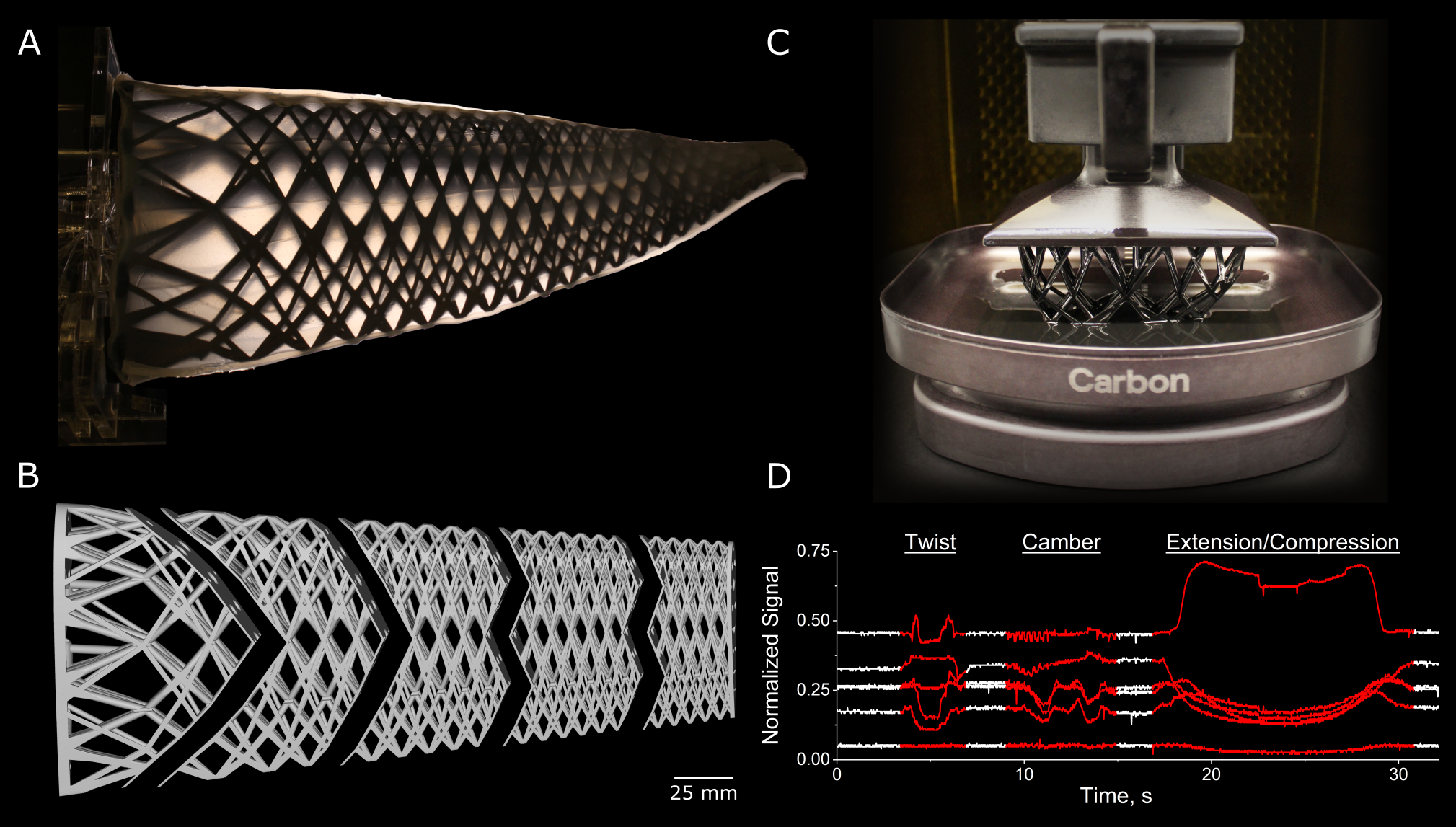}
    \caption{Design Overview. A) Completed wing shown with 60° of upward twist. B) Breakdown of the five segments of the lattice which were assembled post-printing C) DLS 3D-printing of a wing lattice segment on the Carbon M1, using EPU 40 resin. D) Example of sensor measurements across each morphing mode: the highlighted morphing data (red) indicates the difference against the static baseline (white)\label{overview}}
\end{figure*}
\section{Introduction}
Conventional aircraft wings are highly optimized, yet typically only for a single flight profile (e.g., cruise), compromising on other regimes and making up for this through geometry change with discrete control surfaces. In contrast, morphing wings use continually variable geometry adaptation to smoothly vary the aerodynamic properties of the wing, maintaining optimality over a wide regime of flight conditions. This results in a wing with higher aerodynamic efficiency, reducing such adverse effects as flow separation and parasitic drag, providing the potential for future aircraft to push the boundaries of the flight envelope \cite{Barbarino2011}.

Due to these advantages, morphing wings have consistently been of interest in aeronautics research and development, with many wings and adaptive control surfaces having been developed and flight-proven in recent years. The history of these wings stretches back to the Wright Brothers, who applied "wing warping" in their original Wright flyer, twisting the wing via a series of pulleys and cables. Since then, a number of fighter-jets have made use of variable-sweep morphing --- the Bell X1 being the first, followed by examples such as the F-14 Tomcat and the Rockwell B1 Lancer. Further jets applied a combination of morphing modes: the Mission Adaptive Wing (MAW) achieved both sweep and wing camber morphing when it was implemented on the AFTI-F111; and the Active Aeroelastic Wing from the X-5 program made use of twist and camber morphing \cite{Weisshaar2013,Perry1966,MAW,x53,Kress1983}.

However, many of the morphing benefits seen by these existing jets do not easily translate to aircraft of all scales --- for instance, variable-sweep morphing aids with managing drag in the transonic/supersonic region, and provides little aid to slower-speed aircraft \cite{Weisshaar2013,Kress1983}. And, the current state of morphing wings outside of military applications suffers from actuation limitations - piezoelectrics are limited to very small ranges of strain (0.1\%) \cite{Giurgiutiu2000}, and shape memory alloys (SMAs) are often limited by their fatigue life and slow actuation response \cite{Barbarino2014}.

Therefore, an opportunity exists for developing effective morphing wings to provide benefits to a broader range of aircraft types. Notably, recent research has highlighted the advantages of distributed compliance and adaptive compliant structure design. Kota et. al. and Flexsys have developed FlexFoil, a flight-tested morphing wing surface with high-performance camber and twist control \cite{Kota2016}. Additionally, Cramer and Jenett et. al. demonstrated spatially programmable anisotropy via composite cellular lattice structures, in two separate compliant and lightweight wings \cite{Cramer2019,Jenett2017}. Bioinspired design can also lead to novel morphing wing architectures at a smaller-scale, as seen in pigeon-based robotic wing research \cite{Chang2020}. Furthermore, previous work in soft lattice structures and embedded fiber-optic sensing was a key inspiration for this paper, which provided an example of variable compliance within a cylindrical lattice volume, and demonstrated the use of the optical sensor network for state estimation \cite{opticalLace}.

In this work, we introduce a new framework for the design of morphing wings, which we denote \textit{Autonomous Material Composites} --- continuous material composites with co-integrated structure, actuation, and sensing designed to interpret and respond to the environment. We develop a 25 cm long wing, using two primary techniques for designing 3D-printed compliant lattice structures: (1) warping the dimensions of a lattice unit cell over the extent of the structure and (2) functionally-grading the strut thickness. Within this wing, we also integrate a network of optical strain sensors (Optical Lace; OL) for detecting the structural deformation. Our actuation system, which enables the morphing, is capable of three degrees of freedom (DOF), producing high deformations with no buckling of the structure or the skin. Testing the sensor network reveals good sensitivity to each morphing mode and distinguishable trends in signal amplitude under deformation, which is essential for state estimation. Furthermore, testing the geometries associated with each morphing mode in a low-Reynolds wind tunnel, we show that this morphing does lead to the desired changes in aerodynamic properties, such as the lift-to-drag ratio. An overview of this system can be found in Figure \ref{overview}. 

\begin{figure*}[t]
    \includegraphics[width=\textwidth]{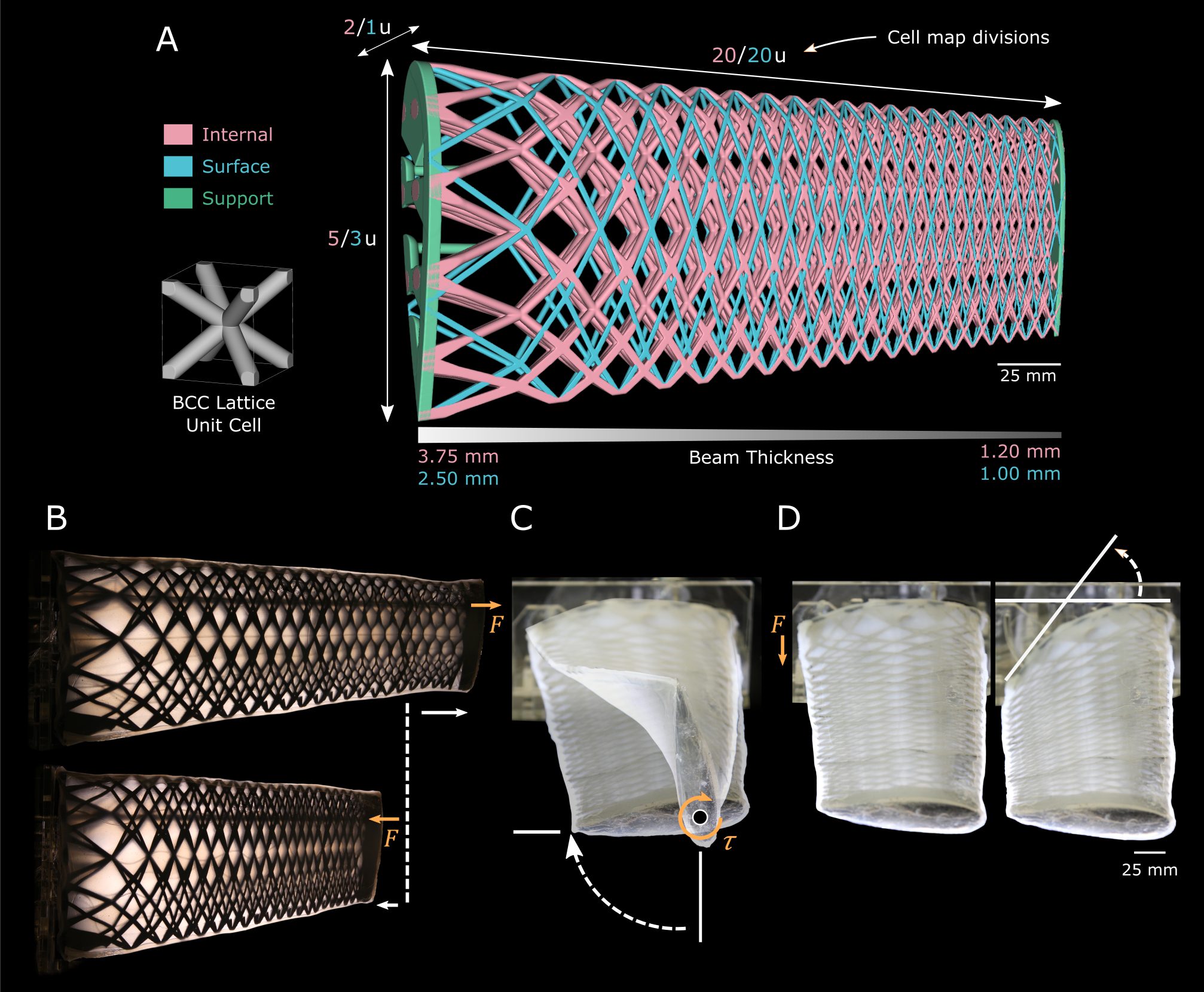}
    \caption{Lattice design and morphing capability. A) nTopology CAD model showing the merged internal and surface lattice structures, along with key design parameters. B) Extension/compression morphing. C) Twist morphing. D) Camber morphing.\label{design}}
\end{figure*}
\section{System Design}
\subsection{Lattice} 
Enabling the tri-modal morphing capability of this wing required careful design of the structure to handle the challenge of spatially-varying mechanical properties. The key engineering contradiction to solve is enabling buckle-free compliance in each morphing DOF, while maintaining sufficient stiffness in the structure to support the wing's weight and expected aerodynamic loading. By using 3D-printed elastomeric lattice structures with spatially-variable warping of the unit cell bounding boxes and functional grading of the beam thicknesses, we tuned the wing's mechanical properties to allow for each region of the wing to morph to the desired geometry. At the root, the wing is resistant to spanwise bending yet compliant in camber; at the tip, the wing is stiffer in camber, yet much more compliant in spanwise extension and twist. Therefore, when connected to our actuator system, we can localize the camber morphing towards the root, and twist at the tip --- the optimal locations for the resulting changes in the aerodynamic lift profile.

The resulting metamaterial consists of a combination of conformal body centered cubic (BCC) lattice structures. By using cell map warping and functional grading of beam thicknesses, we used this structure to form a wing with variable mechanical properties while maintaining continuity in the lattice structure. The structure is composed of two fused architectures: a main internal lattice to support the primary loading and actuation stresses, and a secondary surface lattice to maintain connection with an elastomeric skin for a smooth interface with the surrounding air flow.

(Note: To design the lattice, we relied heavily on the engineering design software nTopology, which supports all of the concepts mentioned below.)

We designed this morphing lattice based on modulating two factors of the BCC unit cell: the relative sizing of the bounding box in each of its three axial directions (which dictates the directional alignment of the beams), and the beam thickness. We relied on the following principle: Given an axially-aligned local coordinate system for a BCC lattice with basis vectors $u$, $v$, and $w$, a unit cell with an extended length in $u$ will result in increased alignment of the beams in this direction, decreasing the compliance from axial loading in $u$ at the cost of increased compliance (axial and bending moments) in $v$ and $w$. Further, considering beam theory principles for a round beam, the flexural rigidity is proportional to the radius to the fourth power (Eqn. \ref{rigidity}), and strain is inversely proportional to the square of the radius (Eqn. \ref{strain}). Therefore, small changes in the radius can dramatically affect the deformation of the lattice. 

\begin{equation}\label{rigidity}
    \text{Flexural Rigidity} = E\frac{\pi r^{4}}{4}
\end{equation}

\begin{equation}\label{strain}
    \varepsilon = \frac{F}{\pi r^{2} E}
\end{equation}

To tune these unit cells, we began with an initial segmentation of 3D Euclidean space with the desired number of cells, and then used cell mapping to map the unit cell boundaries to our design space (the boundary of the wing). Since the wing is non-rectangular, the cell map must conform to the curvature of our design space, so we built this mapping between the upper and lower surfaces of the wing, via an imported CAD model. Refer to Figure \ref{design}A for the specific ($u$, $v$, $w$) divisions used for the internal and surface lattices. 

With this initial mapping created, warping the map affects the relative sizing of cells in each direction based on a scaling vector field. We applied a linearly variable gradient field such that the spanwise size of the unit cells is increased at the root, and decreased at the tip of the wing.

After generating our BCC lattice structure within the cell map boundaries, the beams exist in an initial configuration as a network of nodes and edges. We then thicken the beams based on an offset parameter from an edge (e.g., the beam radius), determined via a continuous field or function within 3D space. 

We applied this warping and beam thickening technique for variable compliance in the wing. At the root, higher beam thickness with an extended unit cell dimension along the span allows for compliance in camber, yet rigidity against bending loads from the weight of the wing and aerodynamic loading. Then, at the tip of the wing, the narrow cell dimension along the span and reduced beam thickness increases compliance for extension and improves the range of twist.

For both the thickness and warping fields, a tangent continuity in the spanwise gradient aids in the local compliance differences between the root and the tip regions. By reducing these gradients at either end of the wing, we better localize the desired mechanical properties within the optimal regions for aerodynamic control (camber and increased lift at the root; twist and roll control towards the wingtip) while maintaining continuity in the structure. Additionally, through this process of tailoring the compliance in the morphed regions, we reduce the necessary stress required to actuate the structure, and reduce the risk of buckling at the surface. 

While we have described the design of the solid region of the lattice, the vacant regions also had to be co-designed to allow for integrating the actuation and sensing components. One of the most critical components of this system is a steel actuation rod, which is routed through the aerodynamic center of the wing and extends over the full length of the span. While a number of candidate lattice designs can satisfy the spatially-variable compliance target, few can also satisfy this actuator position requirement without interfering with the lattice continuity. Therefore, our combination of a BCC unit cell with the specific chordwise division of cells (Figure \ref{design}A) allows for this continuously-empty region of the lattice and the unhindered assembly of the actuation system. In addition, our sensing OL fibers were routed through the lattice in 3D loops, from the root to the tip and back again, with narrow channels included to hold these sensors in place even under high deformations. 

To finalize the wing, the surface and interior lattices are merged via Boolean union operations, along with all additional, non-lattice supporting elements in the wing. Additional Boolean intersection operations are used to divide the lattice into five separate components for manufacturability, with the boundaries between these components aligned with the beams in the lattice to minimize their impact on the structure (Figure \ref{overview}B) (see the Fabrication section for the reasoning behind this). After determining the routing pathways for the optical lace, actuators, and other elements in the wing, any material that would interfere with these paths is removed with a final Boolean subtraction operation. Finally, the parts are exported as STLs for printing.

\begin{table}
\small\sf\centering
\caption{\label{tab:params}Wing geometry parameters}
\begin{tabular}{ll}
\toprule
Airfoil & NACA 0020 \\
Taper ratio & 60\% \\
Sweep angle & 0° \\
Span & 250 mm \\
Chord length (root) & 130 mm \\
\bottomrule
\end{tabular}
\end{table}

\begin{figure}[t]
    \includegraphics[width=\columnwidth]{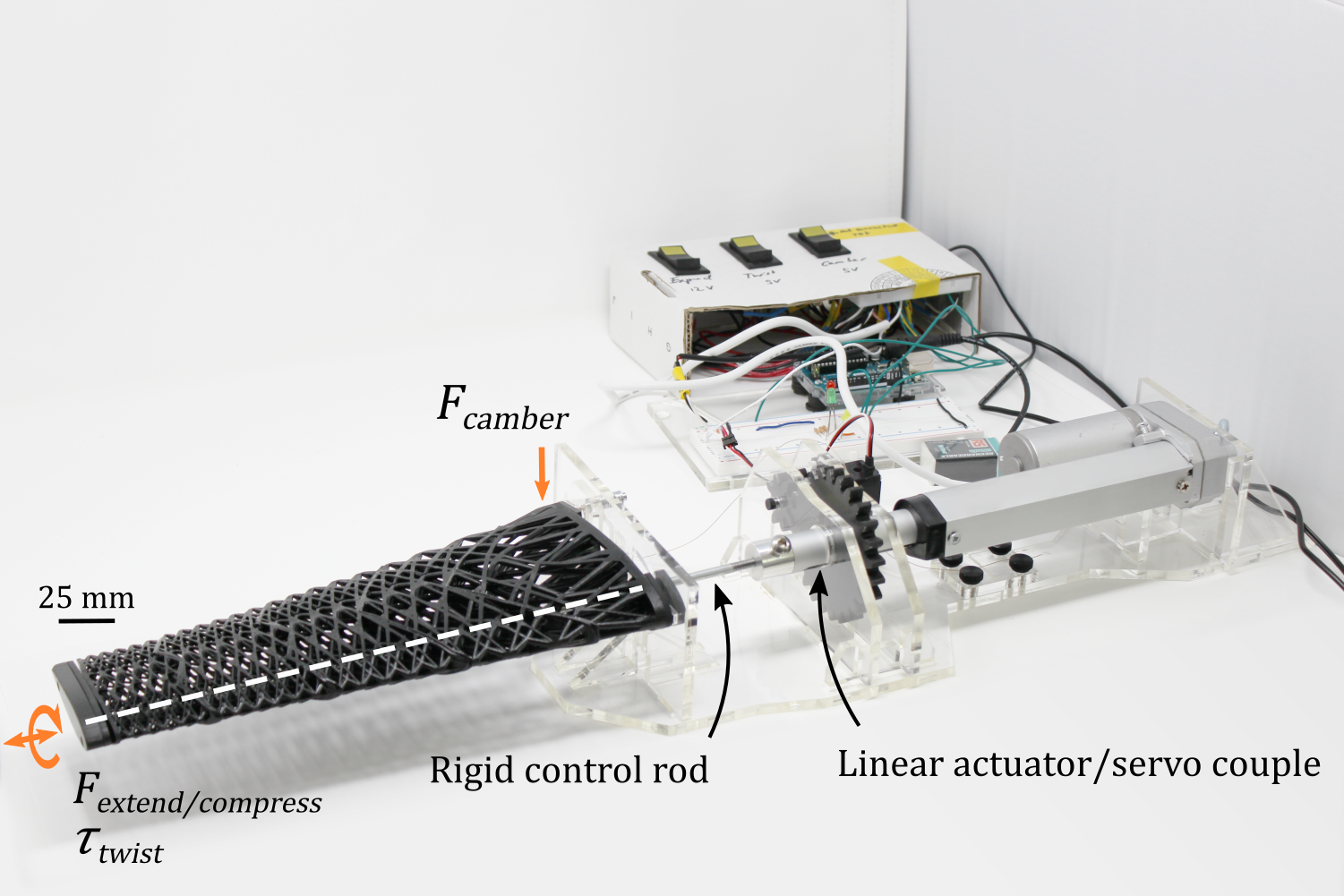}
    \caption{Actuator testing setup\label{actuation}}
\end{figure}  
\subsection{Actuation} 
The wing can support actuation (morphing) in three independent degrees-of-freedom: twist, camber, and extension/compression (Figure \ref{design} B, C, D). This combination of DOFs allows for a wide range of theoretical benefits to the aerodynamic performance, including the following three examples: (1) Adjusting the twist of the wing not only provides roll control, but also the capability to adjust the spanwise lift distribution; (2) Modifying chordwise camber redirects the flow at the trailing edge of the wing to adjust the lift coefficient and maintain flow attachment through various flight speeds and conditions; (3) An expanded span will allow for improved roll control by extending the effective moment arm, and compression can allow for increased maneuverability in densely occluded environments. Combining these morphing effects can also be beneficial --- for instance, active control of both twist and extension/compression can mitigate flutter by adjusting the loading distribution and vibrational modes \cite{Anderson1989}.

To support this morphing, we routed a steel rod through the quarter-chord point of the lattice, stretching outwards from a linear actuator at the root and terminating at a rigid structure at the tip of the wing. This rod delivers both the twisting moment and extension/compression force. To do this, we milled a slot into the linear actuator rod and removed the rotation lock on the leadscrew nut, so that we can externally control the rotation of the rod with a servo motor. The servo is connected to a pin which slides freely when the actuator is purely extending/retracting, but which can also deliver the torque to turn the rod (and thus twist the wing). At the root, adjustable cable tension via our mount produces the downward force necessary to camber the wing. A view of the testing setup is available in Figure \ref{actuation}. 

Given the elastomeric nature of the lattice, the wing can sustain a wide range of morphing with minimal actuation force and no plastic deformation. Morphing via twist has an operating range of ±90°, camber can redirect the direction of the trailing edge downwards by up to 60°, and extension/compression can handle +30 mm / -20 mm (+12\% / -8\%). This range of extension/compression is currently limited by the stroke length of the linear actuator; a wider range of motion was tested outside of the actuator setup up to +70 mm / -57 mm (+28.0\% / -22.8\%). Buckling effects become visible at this lower bound of -22.8\%, but notably, the steel rod routed through the structure significantly delays the onset of this buckling, because it constrains the motion of the inner lattice structure along its axis. 

Extension and compression require the most force from the actuators: 9.29 and -8.48 N for the +28.0\% / -22.8\% extension and compression cases, respectively. Twist and camber were each held at their maximum positions under no aerodynamic load with 0.8 Nm torque. 

\subsection{Sensors}
Because the elastomeric lattice allows for stress to be continuously propagated, as opposed to being concentrated at a few mechanical joints in traditional morphing approaches, we had the opportunity to embed soft strain gauges throughout the volume of our metamaterial to map the resulting strain evolution. These strain gauges, called Optical Lace (OL) \cite{opticalLace}, are made of fibers of elastomer that propogate light as a waveguide. Upon stretching or bending, they propagate fewer rays, thereby decreasing the signal. With these sensors connected to a microcontroller and monitor, when the wing is morphed, we can view the local deformation and signal throughput in real-time. 

Within the wing, we embedded six optical threads (diameter, d = 0.8 mm; Crystal Tec), looped between the root and tip of the wing such that both ends of the thread terminate at a circuit board located behind the root. This circuit board contains six light emitting diodes ($\lambda \approx$ 875 nm; TSHA4400) and six silicon NPN phototransistors (450 $\leq \lambda \leq$ 1080 nm; BPW85C) that interface with the OLs at a friction-fit connection component. Infrared (IR) light from the LEDs at one end of each optical lace passes through the sensor and terminates at the phototransistor on the opposite side, registering a signal on the corresponding microcontroller (Arduino Mega) port. The light intensity as read by the phototransistors decreases due to stretching or bending of the sensor, and the corresponding deformation in the structure can be observed through the varying amplitude of light passing through the six sensors. 

Figure \ref{sensors}A describes the integration of the six OL sensors. Threading in this pattern allows for the sensors to pass through the lattice unimpeded by the solid actuation components, while also looping around a rigid polyurethane component at the wingtip to reduce shearing stresses on the elastomeric lattice when the sensors are under tension. 

\subsection{Skin} 
The skin material must be able to maintain a smooth surface profile during all classes and magnitudes of morphing, and prevent any discontinuities (i.e. from buckling of the skin) that would significantly decrease the lift-to-drag ratio and overall efficiency of the wing. The skin, therefore, must also tightly conform to the lattice structure under all morphed geometries and be compliant enough to not prevent the motors from actuating the lattice structure.

Given these requirements, we selected a low elastic modulus silicone (E $\approx$ 150 kPa; Smooth-On, Inc; Dragon Skin 10 Slow) \cite{dragonskin} shell to cover the lattice structure with an average thickness of 1.2 mm, a similar material choice as that of other small-scale bioinspired morphing wings with high compliance \cite{batbot}. 

When designing the mold for the skin, we under-sized it by 10\% isotropically, to introduce tension into the skin when wrapped around the wing. This tension allowed the skin to remain tightly attached to the lattice across all of our tests, even when the underlying structure was significantly compressed. 

\section{Fabrication}
\subsection{Lattice: Digital Light Synthesis}
We produced the wing's lattice structure using a a Digital Light Synthesis (DLS) printer (M1; Carbon, Inc.) along with a commercially available elastomeric polyurethane (EPU 40; Carbon, Inc.). DLS, similar to stereolithographic (SLA) printing, relies on the use of light to polymerize and crosslink resilient and tough elastomers at high resolution ($\approx$ 30 $\mu $m feature sizes), rapidly ($\approx$ 20 mm/hr). The printed EPU has a large elongation at break ($\gamma\approx$ 300\%), high tear resistance ($\approx$ 20 kN/m), and moderate elastic modulus (E $\approx$ 8 MPa) \cite{EPU}, meaning the lattice can deform easily, reversibly, and without requiring large motors to induce strain. Unfortunately, due to the thin lattices we designed and the low elastic modulus of the material, printing the entire wing resulted in sagging of the structure. However, we circumvented this issue by printing the lattice in multiple sections (Figure \ref{overview}B) and bonding them post printing. 

Because the EPU 40 material uses orthogonal chemistry, the photopolymerized structure requires a thermal post-cure. Therefore, between the printing and curing processes, we were able to apply unreacted EPU 40 as a glue between the five sections. After curing, the structure becomes monolithic and is chemically indistinguishable across the bonded regions. The additional material from adhesion, however, alters the mechanical properties we initially estimated from the unit cell warping of the lattice. To reduce this effect, we designed the wing's five components to interface along the existing beams within the lattice, minimizing any additional modifications to the structure. 

\subsection{Skin: Molding/Curing}
Producing the skin involved a multi-stage curing process, using a two-part negative mold of the wing boundary to contain the silicone resin. This mold was manufactured with a high surface resolution (28 $\mu $m layer slice) via PolyJet printing (Object 30 Scholar; Stratasys) \cite{Stratasys2021}.

The curing process of the silicone required two primary steps --- a preliminary cure of the silicone on either half of the mold, and a final cure to bond the two halves of the skin together into a hollow shell. We divided 70 grams of Dragon Skin 10 resin between the two halves of the mold, retaining 15 grams for joining the halves. An eight-minute bake in an 80°C oven completed the preliminary cure, and we then applied the retained silicone to the boundaries of the mold and any visible holes. An additional eight minutes in the oven finished the curing process, and once fully cooled, we carefully pried apart the mold to remove the skin. 

With the hollow skin completed, we made an incision at the root of the wing to slide the lattice inside, and then permanently bonded the silicone to the root. This bonding process required two steps: a pretreating of the bonding surfaces on the EPU lattice with Loctite SF 770 primer, and then forming the connection with Loctite 401 instant adhesive applied to the silicone. 

\begin{figure*}[t]
    \includegraphics[width=\textwidth]{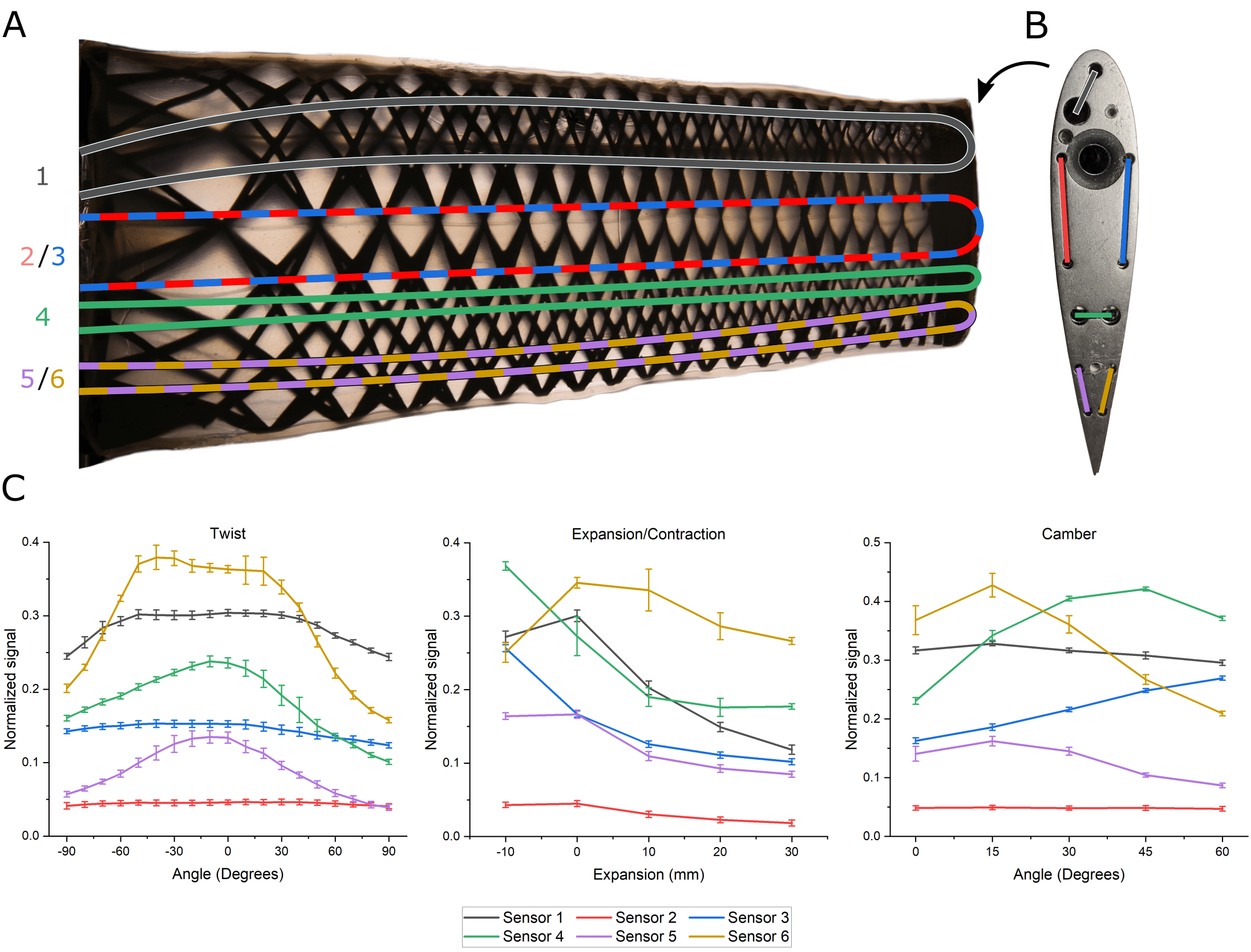}
    \caption{Sensor testing results. A) Locations of the sensors in the wing. B) Additional view of the sensor locations at their connection to the wingtip. C) Signal amplitudes for each sensor across the range of each morphing mode.\label{sensors}}
\end{figure*}  
\section{Sensor Evaluation}
\subsection{Testing Procedure}
Using the open-loop actuator setup, we connected the optical lace and recorded their output to directly compare the light amplitude across each morphing mode and combination of modes. These different wing positions strain the sensors in varying amounts, with the relative signal amplitudes and trends being critical for reliable state estimation. 

For all tests, the other morphing modes not being tested were pre-set to the desired position and held constant through the trials. This typically consisted of pre-setting the extension amount and then testing twist and camber, but pre-setting twist and camber was also evaluated. 

To test the sensor performance under twist, we began data collection with the wing initially set to 0° of twist, and then twisted the wingtip upwards or downwards in 10-degree increments, pausing for three seconds at each increment to allow the sensor reading to stabilize. At the maximum twist (+/- 90°), we twisted the wing in the opposite direction, following the same 10°-and-pause procedure until the wing returned to the 0° position. 

For extension testing, we began the data collection at -10 mm of compression, and then expanded the linear actuator in 10 mm increments. At least three seconds of data was collected at each increment, and then, at the +30 mm extension point, this process was reversed until the wing was back to full compression. 

For camber testing, we followed a similar process to that used for the other tests, where we began data collection with the camber set to 0°, and then gradually increased this in 15° increments until the maximum camber of 60°. This camber was then reduced in the same 15° increments until it returned to 0°. The three-second pauses at each increment were also maintained for this test.

Note that these tests were all performed outside of the wind tunnel so we could purely observe how the signals vary with the changes to the structure induced by morphing. 

\subsection{Results}
Each morphing mode achieved reliable changes in signal amplitude across the full set of sensors, with some sensors having a higher sensitivity to certain morphing modes, or the degree of morph (Figure \ref{sensors}C). 

With twist, we observed a consistent decrease in signal amplitude with increased twisting actuation --- a result of the higher strain on these sensors at large deformations of the wing. Here, sensors 4 and 5 achieved the best performance. 

With extension/compression, for most sensors, signal amplitude decreased with extension (and stretched sensors), though some small gains were visible when moving from compression to extension. Under compression, two sensors (1 and 6) have more slack, so when this additional curvature was removed, the amplitude increased.  

For camber, sensor 3 achieved excellent readings --- nearly linear, and with small standard deviations, while other sensors showed interesting trends of increasing and decreasing depending on when tension was added or removed (depending primarily on if the sensor was positioned on the top or bottom of face of the wing). 

Having certain sensors primarily sensitive to one morphing mode is beneficial for identifying the current state of the wing. For instance, sensor 3 showed low sensitivity to twist, but excellent performance under camber, allowing for easier differentiation between these morphing modes. In general, having multiple sensors with different performance and sensitivity across the morphing domains can account for poor performance in any single sensor (for instance, sensor 2).

Overall, the sensors displayed differentiable trends over each morphing mode. Therefore, combining these trends across the full network of sensors with machine learning models can lead to robust state estimation, as shown by Van Meerbeek and Xu \cite{VanMeerbeek2018,opticalLace}. 

\begin{figure*}[t]
    \includegraphics[width=\textwidth]{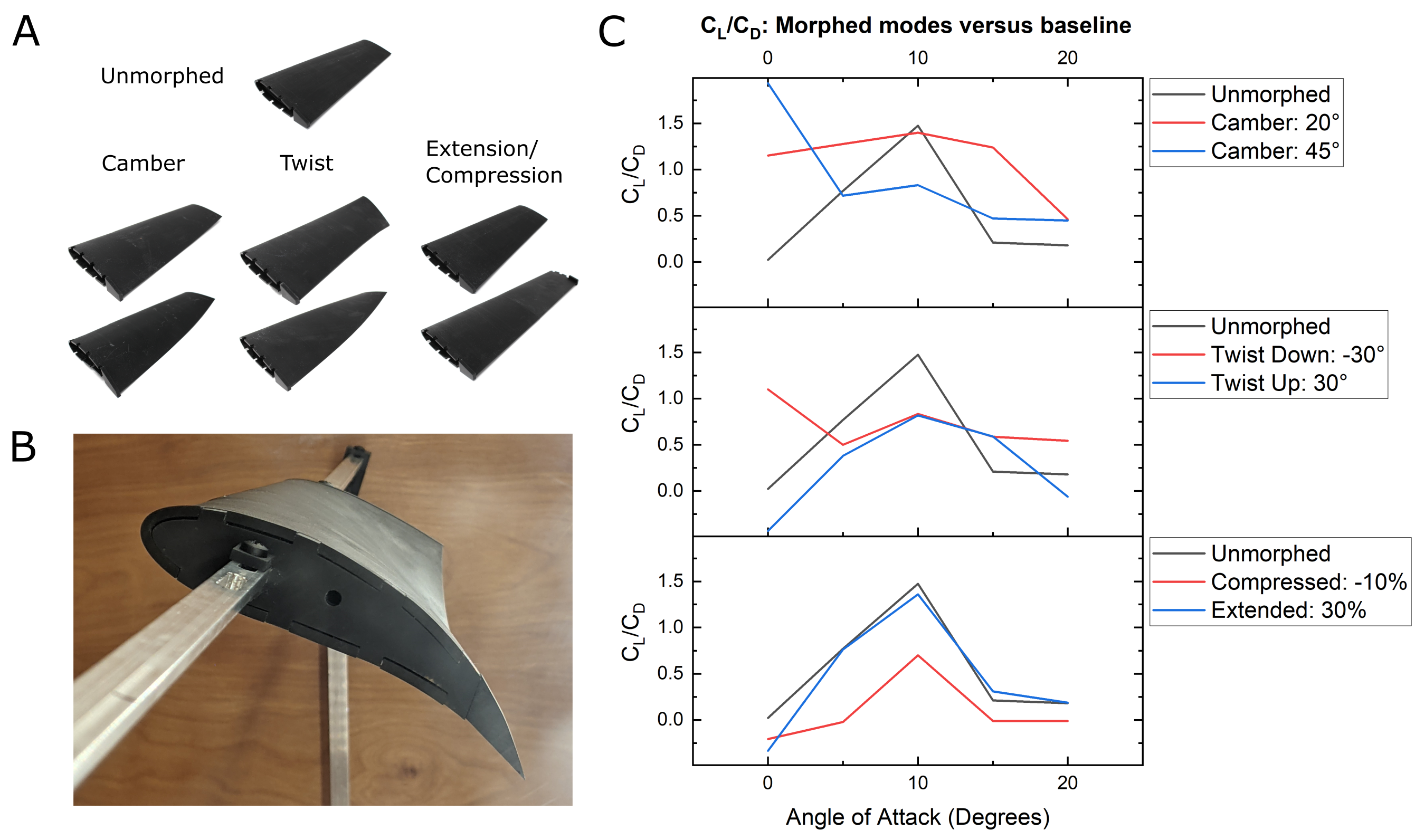}
    \caption{Wind tunnel testing results. A) The seven testing components, representing the unmorphed and morphed geometries across all configurations. B) Example of a model mounted to the testing rig. Shown: 20° camber at $\alpha$ = 20°. C) Coefficient of lift-to-drag comparison for each geometry at various angles of attack.\label{F3 wind tunnel}}
\end{figure*}
\section{Wind Tunnel Evaluation}
\subsection{Overview}
Wind-tunnel testing of the morphed wing geometries allowed for direct comparison of the lift and drag characteristics to evaluate the effectiveness of each morphed DOF on the desired aerodynamic performance (such as an increased lift-to-drag ratio or induced rolling moments).

We produced seven urethane methacrylate (UMA; Carbon, Inc.) 3D-printed testing models to represent the unmorphed configuration and six morphed geometries: camber (20° and 45°), twist (upwards and downwards at 30° each), extended 30\%, and compressed 10\% (Figure \ref{F3 wind tunnel}A). These were attached to the center of a custom-built testing rig for collecting lift and drag force data over multiple angles of attack (0°, 5°, 10°, 15°, and 20°) (Figure \ref{F3 wind tunnel}B). This data was recorded over four straight bar aluminum load cells (TAL220B) with up to 50 N measurement capability each. Smoke tests with high-speed video were also used to confirm the onset of flow separation for the models, which was in line with the data seen from the force gauges. All tests were performed in a custom-built open-loop circuit wind tunnel \cite{windTunnel}.

Note that a drawback of our force-testing rig is that it is only compatible with the rigid testing units that we developed. Connecting all of the actuators to the wing in a similar manner to our testing setup (Figure \ref{actuation}) inside the tunnel was not viable, so testing on the flexible wing remains an important consideration for future work, as we do expect that we would see some adverse aeroelastic effects.

\subsection{Procedure}
We first performed a calibration test to evaluate the aerodynamic effect of the testing rig on the experimental data, and then continued through each of the morphed geometries as follows. We began with the rig set to 0 angle of attack, and set the wind speed to the targeted 6.7 m/s. This wind speed corresponds to a Reynolds number of 50,000, based on the mean aerodynamic chord of the testing elements. We began the data collection and then turned on the tunnel, and once we observed steady-state full-speed airflow for at least 10 seconds, we turned off the tunnel. Data collection ended when the flow in the tunnel had slowed below the tunnel's minimum windspeed sensor sensitivity, and then this process was repeated for each additional angle of attack (5°, 10°, 15°, and 20°). 

\subsection{Results}
Morphing the camber of the wing effectively increased the amount of lift produced by the wing over the range of angles-of-attack ($\alpha$) tested. At $\alpha$ = 0° AOA, the 45° camber wing achieved a maximum lift-to-drag coefficient ratio $C_L/C_D$ of 1.93, 68\% greater than the 20° wing (which yielded a $C_L/C_D$ of 1.15). At increasing $\alpha$, the 45° camber wing sees a rapid decrease in $C_L/C_D$, indicating that this degree of camber is only effective at low $\alpha$ - past this, flow detachment and stall occurs. However, the 20° camber wing maintains a high level of $C_L/C_D$ throughout the testing, and though the flow begins to separate at $\alpha$ = 15°, this still exceeds the results shown from the baseline unmorphed wing. Refer to Figure \ref{F3 wind tunnel}C for the relevant graphs for this section.

Adjusting the twist of the wing, as expected, leads to a net positive/negative force on the wing at $\alpha$ = 0°, based on whether the twist was in the downwards or upwards directions. We found that this twist induced an average 0.042 Nm of rolling moment on the wing, assuming that the center of pressure for this lifting force in the $\alpha$ = 0° case was located at 80\% of the span length (which also corresponds with where the twisting effect was localized in the wing geometry). Notably, even across the range of angles of attack tested, we see that the twisted wings are able to maintain a moderate $C_L/C_D$, indicating reliable maneuverability over a wide flight regime. 

With extension/compression, there is minimal change to $C_L/C_D$ for extension, though there is a decrease for the compressed wing. While a higher aspect ratio wing is expected to reduce the effects of induced drag and trailing-edge vortices, we do not see this in the data due to the scale of the wing and the relatively low change in aspect ratio. Thus, we will solely use this morphing capability for improving roll control with an extended wing. Future wind tunnel tests to collect roll moment torques with a combination of morphing modes (e.g. twist and extension) will be a key priority to clarify this capability. 

\section{Conclusion}
In this work, we have presented a novel means of the design of morphing wings via compliant lattice structures, using functional grading and unit-cell warping to tailor the structure's mechanical properties to the locally-desired morphing mode. Our results demonstrate that a structurally-integrated network of optoelectronic strain sensors can detect the deformation induced by each morphing mode, providing a means for future state estimation and control of the wing. Additionally, the results of our wind tunnel testing validate that morphing in the three degrees of freedom (extension/compression, twist, and camber) results in the desired changes in the aerodynamic properties, such as the lift-to-drag coefficient ratio and the rolling moment. Compared with other morphing wings, we achieved high deformations, provided multi-modal morphing capability, and minimized the complexity of the structure/actuation system with a structure that is designed specifically to accommodate the spatially-varying morphing. 

The wing still has multiple limitations though, primarily involving the manufacturing process and scalability of the design. First, the five-component segmentation of the lattice remains necessary given the size of the Carbon M1 printer and the limitations of the EPU material during the fragile printing process, but this interferes with the continuity in the mechanical behavior of the lattice despite efforts to mitigate this effect. The 3D printing process is also difficult for wings of a larger size, and scalability is also limited by the properties of the materials used. EPU and silicone have low stiffness-to-weight ratios, and any larger wings produced from these materials will be too heavy to support their own weight. The low out-of-plane stiffness of the silicone skin will also result in detrimental aerodynamic effects under higher aerodynamic loads, and at higher-Reynolds flight regimes, the elastic behavior of the material will likely result in significant aeroelastic flutter/divergence concerns. 

Despite these limitations, we remain confident in the potential of this wing, as well as the broadly-applicable potential of Autonomous Material Composites. The embedded sensor network in the wing establishes a platform for developing a novel state estimation and control system in future work. Similar work on state estimation of soft structures with optical lace sensors has been done previously \cite{opticalLace,VanMeerbeek2018}, but the inclusion of this state estimation in the feedback loop of a wing control system is an exciting prospect. With the capability of understanding its own shape and position, the wing will additionally be able to quickly sense and respond to any onset of aeroelastic flutter/divergence, with active control to mitigate these effects. Preliminary work on collecting data for implementing a learning-based sensor model is in progress, with updates to the sensor positioning in the lattice to reduce the measurement standard deviations coming as well. Integration with a small-scale drone will be the final step, with real flight-testing of the wing being invaluable to proving this design as a feasible concept. 

\begin{acks}
The Authors would like to thank nTopology for the trial of their software, Greg Bewley for the use of the BATL wind tunnel, and Joe Sullivan for his help in the machine shop. D.M. especially thanks all ORL members for their continued advice and support.
\end{acks}

\begin{dci}
R.F.S. is a cofounder of Organic Robotics Corporation which licenses intellectual property related to the sensors in this work.
\end{dci}

\begin{funding}
The authors disclosed receipt of the following financial support for the research, authorship, and/or publication of this article: This work was supported by the Air Force Office of Scientific Research, Contract FA9550-20-1-0254.
\end{funding}

\bibliographystyle{SageV}
\bibliography{wingBibliography}

\end{document}